\documentclass{article}

\usepackage[preprint,position]{neurips_2026}

\usepackage[utf8]{inputenc}
\usepackage[T1]{fontenc}
\usepackage{hyperref}
\usepackage{url}
\usepackage{booktabs}
\usepackage{amsfonts}
\usepackage{nicefrac}
\usepackage{microtype}
\usepackage{xcolor}
\usepackage{graphicx}
\usepackage{colortbl}
\usepackage{multirow}
\usepackage{tikz}

\title{Metis AI: The Overlooked Middle Zone Between AI-Native and World-Movers}

\author{
  Xiang Li\\
  Massachusetts General Hospital
}

\begin{document}

\maketitle

\begin{abstract}
The dominant discourse on AI limitations frames the boundary of AI capability as a divide between digital tasks (where AI excels) and physical tasks (where embodiment is required). We argue this framing misses the most consequential boundary: the one \emph{within} digital tasks. We identify a class of tasks we call \textbf{Metis AI}, named for the Greek concept of \emph{metis} (practical, contextual knowledge), that are performed entirely on computers yet resist reliable AI automation. These tasks are not computationally intractable; they are \emph{institutionally}, \emph{socially}, and \emph{normatively} entangled in ways that defeat algorithmic approaches. We distinguish \emph{constitutive metis} (knowledge destroyed by the act of formalization) from \emph{operational metis} (system-specific familiarity that automation can progressively absorb), and propose five structural characteristics that define the Metis AI zone: consequential irreversibility, relational irreducibility, normative open texture, adversarial co-evolution, and accountability anchoring. We ground each in established theory from across the social sciences, philosophy, and humanitarian practice, argue that these characteristics are properties of the tasks themselves rather than limitations of current models, and show that the appropriate design response is not better automation but \emph{centaur architectures} in which humans lead and AI supports.
\end{abstract}

\section{Introduction: The Overlooked Middle}
\label{sec:intro}

Consider three tasks, all performed on computers:

\begin{enumerate}
    \item \textbf{Classifying emails as spam.} A pattern-recognition problem with clear ground truth, abundant training data, and low stakes for individual errors. AI handles this natively.

    \item \textbf{Planning cancer treatment for a 72-year-old patient with stage~III lung cancer, two comorbidities, and a family divided over goals of care.} The oncologist reads the same CT scans and lab results on the same screen as an AI classifier. But the task demands integrating clinical evidence with patient values, navigating family disagreement over aggressive versus palliative care, judging acceptable risk under normative uncertainty, and committing to a treatment course whose consequences: toxicity, quality of life, survival, cannot be undone. AI cannot do this.

    \item \textbf{Performing surgery on a human heart.} This requires physical embodiment: hands, instruments, haptic feedback. AI cannot do this (at least for now) either, but for an entirely different reason.
\end{enumerate}

Tasks~1 and~3 are well understood. Task~1 belongs to what we call \textbf{AI-Native} territory: well-structured digital tasks with clear success metrics where AI systems reliably match or exceed human performance: if it's not yet, eventually AI will be. Task~3 belongs to the \textbf{World-Movers}: tasks requiring physical intervention in the material world, where embodiment, irreversibility of physical action, and the reality gap between simulation and the real world present fundamental barriers, although such barrier is diminishing. The limitations of AI in the World-Mover domain have been extensively studied under the rubric of Moravec's paradox, the sim-to-real gap, and embodied cognition \citep{moravec1988, brooks1991, pfeifer2006}.

Task~2 is different. It is entirely digital. The oncologist could be reviewing scans, reading pathology reports, and conferencing with the patient's family all on a screen. No physical manipulation, no embodiment, no reality gap. And yet it resists AI automation as stubbornly as surgery does. Why?

This paper argues that the most consequential and under-theorized boundary in AI capability is not between digital and physical tasks, but \emph{within} digital tasks themselves. Between the spam classifier and the oncologist lies a vast middle zone of tasks that are computationally accessible but institutionally, socially, and/or normatively resistant to automation. Here by \textbf{institutions} we mean the durable structures of law, professional practice, governance, and social norms that organize human action: not individual organizations but the rules and expectations within which organizations operate. We call this zone \textbf{Metis AI}---named for the Greek concept of \emph{metis}, the practical, contextual, relational knowledge that these tasks demand and that formalization destroys.

Our contribution here is to characterize this zone through five structural properties, not of current AI systems, but of the tasks themselves and the human institutions in which they are embedded. We draw on theory from social science, philosophy, communication studies, and humanitarian practice to argue that these properties are durable: they will not yield to larger models, more data, or better architectures, because they arise from the structure of human social life, not from computational constraints, but is, and will be a central component of AI for a long time.

\section{Three Zones of AI Capability}
\label{sec:zones}

First of all, We formally propose a simple tripartite framework for the landscape of tasks:

\begin{table}[h]
\centering
\caption{Three zones of AI capability.}
\label{tab:zones}
\begin{tabular}{@{}lllll@{}}
\toprule
Zone & Name & Medium & AI Status & Defining Property \\
\midrule
1 & AI-Native & Digital & AI performs reliably & Well-structured; clear metrics \\
2 & Metis AI & Digital & AI fails despite digital medium & Entangled with institutions \\
3 & World-Movers & Physical & AI requires embodiment & Requires physical intervention \\
\bottomrule
\end{tabular}
\end{table}

\textbf{Zone~1 (AI-Native)} encompasses tasks like text classification, code generation, image recognition, recommendation, and translation. These tasks have been extensively benchmarked, and AI systems achieve human-level or superhuman performance on many of them \citep{liang2023}. They share a common structure (one of many of the following): the problem can be fully specified in information space, success is measurable against a well-defined ground truth, and abundant digital training data exists.

\textbf{Zone~3 (World-Movers)} encompasses surgery, construction, caregiving, and other tasks requiring physical manipulation of the material world. The limitations here are well understood and discussed: Moravec's paradox, the reality gap, and the 120,000$\times$ data gap between robotics and language models \citep{firoozi2025foundation}. We do not address Zone~3 further; it has an extensive literature.

\textbf{Zone~2 (Metis AI)} is our focus. These tasks are performed entirely on computers---via email, documents, spreadsheets, video calls, databases, messaging platforms. The information is digital. The interfaces are digital. And yet the tasks resist reliable AI automation. What do these tasks have in common? They are all computationally accessible. An AI system can read the relevant documents, process the data, generate analyses. But they all resist reliable automation. The bottleneck is not computation. It is something else.

\section{The Master Frame: Techne Meets Metis}
\label{sec:frame}

To understand what makes Metis AI tasks resistant, we draw on a distinction with roots in ancient Greek philosophy and modern social science.

\textbf{Techne} is the formal, universal, codifiable knowledge, the kind that can be written down in manuals, encoded in algorithms, and transmitted without loss to any competent practitioner. AI systems are techne machines. They operate through learned procedures applied to formalized inputs to produce optimized outputs.

\textbf{Metis} is practical, contextual, relational knowledge that resists formalization. \citet{scott1998} gave this contrast its sharpest modern formulation in \emph{Seeing Like a State}: metis is ``the kind of knowledge that can come only from practical experience,'' and high-modernist schemes that attempt to replace it with techne, whether through Soviet central planning, forced villagization, or algorithmic optimization, produce catastrophic failures. Aristotle identified the same distinction as \emph{phronesis}: practical wisdom that cannot be reduced to general rules. Works in \citet{bourdieu1977} called it \emph{habitus}: the tacit, socially acquired dispositions that govern practice without being (and possibly not able to be) consciously articulated.

The techne/metis distinction has been productively applied in software engineering, where \citet{nolan2021} argues that knowledge of a specific production system is metis: acquired through on-call experience and not fully capturable in runbooks. This is an important observation, but it describes what we call \textbf{operational metis}: system-specific familiarity that is difficult to transfer but in principle acquirable by any competent practitioner given sufficient time/effort. Infrastructure standardization like Kubernetes, service meshes, observability platforms will progressively reduce the operational metis burden, much as ``dredging the shipping channel reduces the amount of local piloting knowledge needed'' \citep{nolan2021}.

The metis that defines our Zone~2 is categorically different. We call it \textbf{constitutive metis}: knowledge that is not merely difficult to formalize but that \emph{is destroyed by the act of formalization itself}. When an oncologist sits with a patient and family to plan treatment, the relevant knowledge: how much this patient truly understands, what ``quality of life'' means to someone who has not spoken it aloud, whether the daughter's insistence on aggressive treatment reflects the patient's wishes or her own grief, does not exist prior to and independent of the clinical encounter. It is \emph{constituted} in the conversation and probably beyond that. It cannot be extracted, encoded, and reapplied, because it is not a representation of a state of affairs; it is a mode of engagement with a situation that is itself shaped by the engagement. This is the difference between a pilot who knows a particular harbor and a physician who knows a particular patient: the harbor does not change its channels in response to the pilot's knowledge of them; the patient does.

Three further properties distinguish constitutive metis from operational metis:

\begin{enumerate}
    \item \textbf{Reflexivity:} In constitutive metis, the knower and the known are entangled. The physician's reading of the patient changes the patient's understanding of their own illness. The regulator's interpretation of a statute changes the meaning of the statute. This is Soro's work (\citeyear{soros1987}) reflexivity applied beyond financial markets to all domains where the model and the modeled co-evolve.

    \item \textbf{Normative irreducibility:} Operational metis involves facts about a system (this database fails under this load pattern). Constitutive metis involves judgments about values (what does ``fairness'' require here? what does ``reasonable care'' mean in this case?). These are not empirical questions with discoverable answers but normative questions whose resolution requires human deliberation and legitimation.

    \item \textbf{Accountability constitution:} Operational metis can in principle be delegated: a sufficiently detailed runbook, given enough time, could substitute for experience. Constitutive metis cannot be delegated because the act of bearing personal responsibility for the judgment \emph{is part of what makes it a judgment}. A diagnosis signed by a physician and the same diagnosis produced by an algorithm are not the same social act, even if their informational content is identical.
\end{enumerate}

The claim of this paper is that \textbf{Metis AI tasks are tasks that demand constitutive metis, not techne or operational metis.} They cannot be solved by formalization, no matter how sophisticated, because their essential difficulty lies in properties that formalization destroys. AI can progressively absorb operational metis by learning system-specific patterns, reducing the need for human familiarity. But constitutive metis is not on a continuum with techne; it is a different kind of thing.

The work in \citet{weick1979} distinguished between \textbf{uncertainty} (the problem is missing information) and \textbf{equivocality} (the problem is multiple conflicting interpretations of the same information). AI is powerful at reducing uncertainty: given more data and better models, it finds the signal. But Metis AI tasks are characterized by equivocality. The participants are not ignorant; they disagree about what the situation \emph{means}. More data does not resolve equivocality. Only collective human sensemaking like negotiation of meaning among parties with different perspectives, interests, and stakes can reduce it.

This equivocality/uncertainty distinction provides the cleanest single boundary for whether a task is AI-Native or Metis AI. If the bottleneck is missing information, AI can help. If the bottleneck is conflicting interpretations, AI cannot.

\section{Five Pillars of Metis AI}
\label{sec:pillars}

We further identify five structural characteristics that make digital tasks resistant to AI automation. Each is independently grounded in established theory in communication, social science, humanity, etc. and is mutually irreducible. We provide a distinguishing test for each, demonstrating that it cannot be collapsed into the others. The most durably AI-resistant tasks exhibit three or more simultaneously.

\subsection{Consequential Irreversibility}

\textbf{Definition:} The digital action produces consequences: legal, financial, reputational, institutional, that propagate through human systems and cannot be undone.

A judge publishes a sentence. A trader executes a position. An oncologist commits a patient to a chemotherapy protocol. A content moderator removes a post during a political crisis. In each case, the action is digital (e.g., keystrokes on a screen) but the consequences propagate irreversibly through institutional systems.

AI systems optimize within defined loss functions. This works when errors are recoverable. But when the cost of error is unbounded and unrecoverable, optimization is the wrong paradigm entirely. Several related works also converge on this point.

\begin{itemize}
\item \citet{dixit1994} generalized this logic into a theory of decision-making under irreversibility: any consequential commitment that is irreversible, uncertain in outcome, and discretionary in timing is analytically equivalent to exercising a financial option. The act of committing ``kills'' the option- will permanently destroy the decision-maker's flexibility. Rationality under irreversibility therefore demands a substantially elevated threshold for action, one that reflects not the expected outcome but the value of what is permanently forfeited by committing. For AI systems designed to maximize expected utility, this presents an architectural contradiction: the optimization paradigm that makes AI effective for reversible tasks is precisely the paradigm that fails when a single action propagates irreversibly through institutional systems.
\item \citet{arrow1974} formalized a complementary insight from environmental economics: when a decision is irreversible and its consequences uncertain, there exists a positive ``quasi-option value'' in preserving the capacity to act differently as new information emerges. Standard expected-value optimization i.e., the paradigm that undergirds most AI system design will systematically underweight this value because it treats the decision as a one-shot calculation rather than an evolving commitment. The result is a structural bias toward irreversible action: the optimizer commits too early, too aggressively, and without adequate regard for what is foreclosed.
\item \citet{jonas1979} proposed a ``heuristics of fear'': when consequences are irreversible, the rational posture is to give disproportionate weight to catastrophic outcomes.
\end{itemize}
\textbf{Distinguishing test:} A task can exhibit consequential irreversibility without involving any human interaction, normative ambiguity, adversarial dynamics, or formal accountability. A single keystroke executing a \$500~million algorithmic trade involves pure irreversibility with no relational or normative complexity.

\subsection{Relational Irreducibility}

\textbf{Definition:} The task's core difficulty is managing human social dynamics: trust, persuasion, coalition-building, face-work, emotional calibration, that exist \emph{between} persons and cannot be decomposed into individually modelable states.

We call this property \textbf{relational irreducibility}, a term drawn from critical realism \citep{lawson2012} and relational sociology, where it denotes the principle that relational wholes cannot be reduced to their component parts. The following literature reflect this concept as well:

\begin{itemize}
\item Work in \citet{habermas1981} argued that genuine mutual understanding requires four validity claims: comprehensibility, truth, rightfulness, and \emph{sincerity}: that the speaker means what they say. AI systems can approximate the first three through pattern-matching, but they structurally cannot fulfill the sincerity claim because they lack the subjective interiority that grounds it. An AI can \emph{simulate} sincerity; it cannot \emph{be} sincere. This \textbf{sincerity gap} is not a limitation of current models but a consequence of what these systems are.
\item \citet{goffman1967} showed that every social encounter is governed by ``face-work'': the mutual protection and negotiation of each participant's claimed social identity. Face is not a fixed attribute but an emergent property of the interaction itself: it is constituted in the moment, fragile, and dependent on the other party's real-time responses.
\item \citet{lederach2005} argued that peacebuilding requires ``the moral imagination'': the capacity to imagine oneself in a web of relationships that includes one's enemies and to act creatively within that web.
\end{itemize}
\textbf{Distinguishing test:} A task can be relationally irreducible without being irreversible, normatively ambiguous, adversarial, or formally accountable. Building consensus among a leadership team about organizational culture via Slack and Zoom involves low stakes, no formal rules to interpret, no adversary, and no legal liability, but the relational dynamics are the entire challenge that AI itself cannot overcome.

\subsection{Normative Open Texture}
\textbf{Definition:} The task requires applying rules, norms, or values that are inherently vague, evolving, or essentially contested, where the norm does not determine its own application in novel cases.

This is what the legal philosopher \citet{hart1961} called the \textbf{open texture} of law: the ``penumbra of doubt'' surrounding the ``core of settled meaning'' of any legal concept.

\begin{itemize}
\item \citet{wittgenstein1953} showed that rule-following is a social practice, not a mental computation. ``No course of action could be determined by a rule, because every course of action can be made out to accord with the rule.'' The meaning of a rule in a new case is fixed not by the rule's text but by the community's ongoing practice, what Wittgenstein called agreement in ``forms of life.''
\item \citet{gallie1956} identified a class of concepts he called ``essentially contested'': concepts like justice, democracy, fairness, and art whose correct application is permanently and legitimately disputed. AI requires operationalized definitions to function, but essentially contested concepts resist operationalization by their nature.
\item \citet{scott1998} demonstrated the consequences of ignoring this insight. \emph{High-modernist} schemes failed catastrophically because they replaced contextual, practice-based knowledge (metis) with universal, formal categories (techne). AI-driven rule application is structurally high-modernist.
\end{itemize}
\textbf{Distinguishing test:} A task can exhibit normative open texture without irreversibility, relational complexity, adversarial dynamics, or accountability requirements. A single individual interpreting whether their personal project meets ``reasonable progress'' under a grant's vague terms faces a genuine normative challenge with no other persons involved, nothing irreversible at stake, no adversary, and no legal liability.

\subsection{Adversarial Co-Evolution}
\textbf{Definition:} The task environment includes intelligent agents who continuously adapt to exploit the AI's behavior, creating a non-convergent arms race where every solution reshapes the problem.

The formal structure of this problem is captured by the adversarial variant of Goodhart's Law \citep{manheim2019}. Goodhart's Law states that ``when a measure becomes a target, it ceases to be a good measure.'' The adversarial variant involves intentional, adaptive subversion. No fixed metric can capture a target that moves in response to measurement. Several works add to this concept as well:

\begin{itemize}
\item \citet{anderson2001} established that security against adaptive adversaries is fundamentally an economic problem, not merely a technical one: attack is easier than defense, and the asymmetry is structural rather than contingent. Because adversaries adaptively concentrate effort on whichever point a defense leaves weakest, every fixed security configuration contains its own obsolescence---the defender must protect everywhere while the attacker need only succeed once, and can redirect effort the moment any particular vulnerability is patched.
\item \citet{dalvi2004} formalized the core failure mode: standard classifiers assume the data-generating process is independent of the classifier itself, but in adversarial domains like spam filtering, fraud detection and intrusion detection, this assumption is structurally violated. Because rational adversaries adapt their strategies to circumvent whatever classifier is deployed, classifier performance degrades rapidly after deployment. Classification in adversarial settings is not an optimization problem with a stable solution but an ongoing game with no fixed equilibrium.
\item \citet{biggio2018} synthesize a decade of adversarial machine learning research to characterize the ``reactive arms race'' between learning systems and adaptive adversaries: each deployed model creates the selection pressure for its own circumvention, and each defensive patch redirects rather than eliminates adversarial effort. Their central conclusion is that purely reactive approaches (retraining on newly observed attacks) cannot prevent novel evasion strategies, because the adversary's adaptation is not random drift but intelligent, targeted exploitation of whatever invariant the current model relies upon.
\end{itemize}

\textbf{Distinguishing test:} A task can exhibit adversarial co-evolution without irreversibility, relational complexity, normative ambiguity, or accountability requirements. Detecting AI-generated spam on a low-stakes hobbyist forum involves no irreversible consequences, no human relationships at stake, no normative interpretation, and no personal liability. Yet the spammers can and will adapt to every detection method within weeks.

\subsection{Accountability Anchoring}

\textbf{Definition:} The task requires a specific human to bear personal legal, professional, or fiduciary responsibility for the outcome, not as oversight but as the accountable principal.

This characteristic is possibly the most common form of metis AI, and is
categorically different from the other four. It is not about task difficulty, cognitive complexity, or the structure of the problem. It is about \emph{social and legal infrastructure}, about how societies allocate responsibility. A task can be computationally trivial and entirely within AI's capability, yet require a human principal because the institutional architecture of law, professional practice, and organizational governance demands a named, answerable party.

\begin{itemize}
\item \citet{garfinkel1967} provided the deepest grounding for this claim. In his ethnomethodological framework, accountability is not a post-hoc layer added on top of action but is \emph{constitutive} of social action itself. People make their actions ``accountable'': visibly rational, reportable, subject to scrutiny, in the very doing of them.
\item \citet{santoni2021} identified four distinct accountability gaps that arise when AI is inserted into decision chains: culpability gaps, moral accountability gaps, public accountability gaps, and active responsibility gaps. These gaps are structural, arising from the architecture of AI systems, not from insufficient capability.
\end{itemize}

\textbf{Distinguishing test:} A task can require accountability anchoring without irreversibility, relational complexity, normative ambiguity, or adversarial dynamics. Signing a simple, routine tax return is computationally trivial, normatively clear, involves no negotiation, and faces no adversary, but someone must put their name on it under penalty of perjury.

\paragraph{The interplay among the five pillars}
The five pillars are mutually irreducible where each can appear in isolation, as the distinguishing tests demonstrate. But in practice they cluster and amplify one another. Consequential irreversibility and accountability anchoring form the tightest pair: societies rarely tolerate irreversible actions without a named responsible party, so the two almost always co-occur. Normative open texture and adversarial co-evolution create a feedback loop: adversarial actors exploit precisely the penumbra of doubt that vague norms create, and each adaptive exploit forces norm reinterpretation, which opens new ambiguity for the next round. Relational irreducibility intensifies both: norms like ``good faith'' or ``fairness'' are interpreted \emph{within} relationships, not in the abstract, and the accountable party is accountable \emph{to} someone: a patient, a defendant, a community. This clustering explains why the most durably AI-resistant tasks exhibit three or more pillars simultaneously: each additional pillar closes off a different category of automated workaround, and together they form a reinforcing structure that no single architectural improvement can dissolve.

\section{Case Studies}
\label{sec:cases}

We now apply the five-pillar framework to types of Metis AI tasks, demonstrating how the pillars interact in practice.

\subsection{Content Moderation of Politically Contested Speech}
A social media platform must decide whether a viral post shared two million times in four hours constitutes ``incitement to violence'' or ``legitimate political protest.'' The post uses ambiguous metaphorical language in a regional dialect. The political context is a disputed election in a country with a recent history of ethnic violence. An AI classifier trained on prior moderation decisions flags the post with 73\% confidence.

\textbf{Pillar analysis:} Consequential irreversibility is \emph{high}: removing the post silences a political voice; leaving it up risks contributing to incitement. Normative open texture is \emph{high}: ``incitement to violence'' is an essentially contested concept \citep{gallie1956}. Adversarial co-evolution is \emph{high}: political operatives and disinformation networks actively study and exploit moderation policies. Relational irreducibility and accountability anchoring are both \emph{moderate}.

\textbf{Metis/techne diagnostic:} The classifier reduces uncertainty (is this post more like previously flagged posts or approved ones?). But the task is dominated by equivocality: the same words can be read as incitement or protest depending on interpretive frame, and the interpretive frame is itself politically contested.

\subsection{Humanitarian Aid Allocation Under Competing Claims}

A humanitarian organization must allocate limited medical supplies across three refugee camps during a disease outbreak. An optimization algorithm recommends allocating 60\% of supplies to Camp~A (highest infection rate), 30\% to Camp~B, and 10\% to Camp~C.

\textbf{Pillar analysis:} Consequential irreversibility is \emph{high}: supplies sent to one camp cannot be recalled. Normative open texture is \emph{high}: the optimization algorithm maximizes lives saved, but Camp~C contains a persecuted ethnic minority; deprioritizing it may violate the humanitarian principle of impartiality. Accountability anchoring is \emph{high}: under the Accountability to Affected Populations framework, the organization must be answerable to the refugees themselves. Adversarial co-evolution and relational irreducibility are both \emph{moderate}: local governments inflate casualty numbers, and the allocation must be negotiated with camp leaders whose trust was built through years of relationship.

\textbf{Metis/techne diagnostic.} The optimization algorithm operates on \emph{uncertainty}. But the actual decision is dominated by \emph{equivocality}: what does ``need'' mean when it intersects with political manipulation, ethnic persecution, and compromised data?

\subsection{Regulatory Compliance Under Novel Statute}
A financial institution must determine whether its new AI-driven lending product complies with a recently enacted ``algorithmic fairness'' statute that prohibits ``unjustified disparate impact'' on protected groups. The statute provides no definition of ``unjustified,'' no threshold for ``disparate impact,'' and no guidance on which of the thirty-plus proposed fairness metrics to use (since many are mathematically mutually exclusive).
\textbf{Pillar analysis.} Normative open texture is \emph{high}: ``unjustified disparate impact'' is an open-textured legal standard par excellence. \citeauthor{wittgenstein1953}'s rule-following problem applies directly: the statute does not determine its own application. Accountability anchoring is \emph{high}: the Chief Compliance Officer must personally certify compliance. Consequential irreversibility, relational irreducibility, and adversarial co-evolution are all \emph{moderate}.
\textbf{Metis/techne diagnostic.} An AI system can compute dozens of fairness metrics and identify statistical disparities. This reduces \emph{uncertainty}. But the task is dominated by \emph{equivocality}: which metric matters? What counts as ``unjustified''? These questions require legal judgment, institutional knowledge, and political reading: metis that no fairness toolkit can provide.

\section{Domain Analysis: Where Do Major AI Applications Fall?}
\label{sec:domains}

By applying three-zone framework and five-pillars characterization to the major domains where AI is being deployed at scale (finance, healthcare, law, and education), we revealed a consistent pattern: within each domain, some tasks are genuinely AI-native, while others are deep Metis AI, and the industry discourse routinely conflates the two. Such conflation potentially create a misleading impression that success in automating bounded, well-specified tasks implies readiness to automate socially-embedded tasks that require judgment, accountability, and institutional legitimacy.

\subsection{Finance: Is Stock Market Modeling Metis AI?}
Finance illustrates the zone distinction with unusual clarity because different financial tasks span the entire spectrum within a single institution.
\textbf{AI-Native financial tasks:} High-frequency market-making, index-tracking, and options pricing are paradigmatic Zone~1 activities. They operate on well-structured numerical data, optimize against explicit objective functions, and succeed or fail against unambiguous metrics. Algorithmic trading now accounts for roughly 60--70\% of U.S.\ equity volume.
\textbf{Where finance becomes Metis AI:} The boundary appears precisely where \citeauthor{soros1987}'s (\citeyear{soros1987}) \emph{reflexivity} takes hold. Soros argued that financial markets are not passive objects to be modeled but reactive systems that change in response to the models applied to them. This is adversarial co-evolution in its purest financial form: the model reshapes the market, which invalidates the model. \citet{danielsson2013} formalized this as \emph{endogenous risk}: the insight that risk in financial markets is not exogenous data to be measured but is partly created by the risk models themselves. When market participants rely on similar models, they act in concert during crises, creating the very volatility they sought to hedge against.
The Knight Capital disaster (August 2012) and the 2010 Flash Crash demonstrate the consequences. Knight Capital lost \$460~million in 45~minutes when a dormant algorithm reactivated: a case study in consequential irreversibility in tightly coupled systems \citep{perrow1984}. The 2010 Flash Crash erased \$1~trillion in market value in minutes as algorithmic traders entered a ``hot potato'' cascade. Both events were entirely digital, involved no physical embodiment, and yet were profoundly resistant to algorithmic control precisely because the algorithms were participants in the system they were trying to manage.
\textbf{Credit underwriting} sits squarely in the boundary zone. AI can predict default probability with high accuracy which is uncertainty reduction. But fair lending law (ECOA, the Fair Housing Act) requires that lending decisions avoid ``unjustified disparate impact'' on protected groups. \citet{obermeyer2019} demonstrated that a widely used healthcare algorithm exhibited racial bias because it used cost as a proxy for need. The same proxy problem pervades lending: AI models with 1,000+ input variables routinely use features that correlate with race, creating disparate impact that is normatively indeterminate and legally open-textured. Researchers have proven that it is mathematically impossible to satisfy competing definitions of fairness simultaneously, a result that transforms a technical problem into an irreducibly normative one.

\begin{table}[h]
\centering
\caption{Financial tasks across the zone spectrum.}
\label{tab:finance}
\small
\begin{tabular}{@{}lll@{}}
\toprule
Financial task & Zone & Dominant pillars \\
\midrule
High-frequency market-making & 1 (AI-Native) & --- \\
Index tracking / passive strategies & 1 (AI-Native) & --- \\
Fraud detection & 2 (Boundary) & Adversarial, irreversibility \\
Credit underwriting & 2 (Boundary) & Normative, accountability \\
Active fund management in crises & 2 (Boundary) & Adversarial, irreversibility \\
M\&A negotiation & 2 (Boundary) & Relational, irreversibility \\
Regulatory compliance (AML/KYC) & 2 (Boundary) & Normative, adversarial, accountability \\
\bottomrule
\end{tabular}
\end{table}

\subsection{Healthcare: Is Diagnosis Accuracy Everything?}
Healthcare is the domain where the gap between AI's demonstrated capability and its deployable utility is most revealing. Over 1,250 AI-enabled medical devices have been authorized by the FDA as of 2025, the vast majority in radiology. In narrow diagnostic tasks like detecting diabetic retinopathy from retinal scans, classifying skin lesions, identifying pneumothorax on chest X-rays, etc., AI matches or exceeds specialist performance. These are genuinely AI-native tasks.

But diagnosis is not a standalone event. It is a station in a clinical workflow that includes history-taking, shared decision-making, treatment planning, and follow-up. These surrounding activities are deeply Metis AI.

\textbf{The accuracy-is-everything fallacy.} \citet{topol2019} argued in \emph{Deep Medicine} that AI's diagnostic accuracy could free physicians to spend more time with patients. The premise is correct, but the framing obscures a deeper problem: diagnostic accuracy is necessary but not sufficient for clinical value. IBM Watson for Oncology demonstrated this when it was deployed at multiple cancer centers and produced treatment recommendations that oncologists rejected in up to 50\% of cases: not because the recommendations were computationally wrong but because they failed to account for patient comorbidities, local drug availability, cultural treatment preferences, and insurance constraints. Watson could reduce uncertainty (what does the evidence say?); it could not reduce equivocality (what does this evidence mean for \emph{this} patient?).

\textbf{Treatment planning} is Metis AI because it requires integrating clinical evidence with patient values, family dynamics, resource constraints, and normative judgments about acceptable risk. As mentioned in the very beginning, a 75-year-old patient with early-stage cancer may prefer watchful waiting; a 35-year-old with the same diagnosis may want maximum intervention. The ``right'' answer depends on relational knowledge and normative judgment: both irreducible to pattern-matching.

\textbf{End-of-life care planning} represents perhaps the deepest Metis AI task in all of healthcare. The conversation itself: navigating a family's grief, interpreting a patient's values when they can no longer speak for themselves, deciding whether to withdraw life support, is constitutively relational and normatively open-textured. The task demands what Aristotle called \emph{phronesis}: practical wisdom applied in the particular case.

\begin{table}[h]
\centering
\caption{Healthcare tasks across the zone spectrum.}
\label{tab:healthcare}
\small
\begin{tabular}{@{}lll@{}}
\toprule
Healthcare task & Zone & Dominant pillars \\
\midrule
Radiology image classification & 1 (AI-Native) & --- \\
Pathology slide analysis & 1 (AI-Native) & --- \\
Drug interaction checking & 1 $\to$ 2 & Accountability \\
Clinical decision support & 2 (Boundary) & Normative, relational, accountability \\
Psychiatric/therapeutic care & 2 (Boundary) & Relational, accountability \\
Organ transplant allocation & 2 (Boundary) & Normative, accountability, irreversibility \\
End-of-life care planning & 2 (Boundary) & Relational, normative, irreversibility, accountability \\
\bottomrule
\end{tabular}
\end{table}

\subsection{Law: The Gradient from Document Review to Judgment}
The legal profession has been a testing ground for AI since at least 2012, when Judge Andrew Peck approved technology-assisted review (TAR) in \emph{Da Silva Moore v.\ Publicis Groupe}, establishing that predictive coding could replace manual document review in e-discovery. Legal e-discovery is now substantially AI-native.
But the legal domain exhibits a sharp gradient from AI-native document processing to deep Metis AI judicial reasoning.
\textbf{Judicial sentencing and risk assessment:} The COMPAS controversy \citep{angwin2016} is the canonical example of Metis AI in the legal domain. COMPAS, a recidivism risk assessment tool, was shown to exhibit racial disparities: Black defendants were nearly twice as likely as White defendants to be falsely labeled high-risk. Subsequent formal work showed that, when base rates differ across groups and prediction is imperfect, a risk score cannot simultaneously satisfy calibration/predictive parity and equalized error-rate conditions\citep{chouldechova2017,kleinberg2017}. This impossibility result transforms what appeared to be a technical bias problem into an irreducibly normative one: society must \emph{choose} which definition of fairness to prioritize, and that choice cannot be delegated to an algorithm.

The \emph{State v.\ Loomis} (2016) decision illustrates accountability anchoring: the Wisconsin Supreme Court upheld the use of COMPAS scores at sentencing but required judges to exercise independent judgment, acknowledging that algorithmic risk scores cannot bear the accountability that criminal sentencing demands.

\textbf{Legal argumentation and trial strategy} resist automation for relational and adversarial reasons. The \emph{Mata v.\ Avianca} incident (2023), in which lawyers submitted AI-generated briefs containing fabricated case citations, demonstrated a different failure mode: not that AI produced wrong answers, but that its confident simulation of legal reasoning proved indistinguishable from competent legal research, until verification revealed the citations were hallucinated. The UK Post Office Horizon scandal, in which over 900 sub-postmasters were wrongly convicted based on faulty automated accounting software, illustrates the catastrophic consequences of treating Metis AI tasks as AI-native: when an institution trusts algorithmic output as ground truth in a domain that demands human judgment, injustice scales with efficiency.

\begin{table}[h]
\centering
\caption{Legal tasks across the zone spectrum.}
\label{tab:legal}
\small
\begin{tabular}{@{}lll@{}}
\toprule
Legal task & Zone & Dominant pillars \\
\midrule
E-discovery / document review & 1 (AI-Native) & --- \\
Standard contract analysis & 1 $\to$ 2 & --- \\
Legal research & 1 $\to$ 2 & Accountability \\
Regulatory interpretation & 2 (Boundary) & Normative, accountability \\
Sentencing / bail decisions & 2 (Boundary) & Normative, irreversibility, accountability \\
Trial strategy / litigation & 2 (Boundary) & Relational, adversarial \\
Bespoke contract negotiation & 2 (Boundary) & Relational, normative \\
Mediation of complex disputes & 2 (Boundary) & Relational, normative, irreversibility \\
\bottomrule
\end{tabular}
\end{table}

\subsection{Education: The Banking Model Automated}
Education presents perhaps the most philosophically revealing test of the Metis AI framework, because it forces a prior question: \emph{what is education for?}
If education is information transfer, \citeauthor{freire1970}'s (\citeyear{freire1970}) ``banking model,'' where teachers deposit knowledge into passive student-receptacles, then AI tutoring systems are a strict improvement. Intelligent tutoring systems like Khanmigo (Khan Academy + GPT-4) and Duolingo demonstrate genuine AI-native capability in well-structured domains: arithmetic, grammar, vocabulary, standardized test preparation. These tasks belong in Zone~1.

But if education is what Freire called ``problem-posing'': a dialogical process through which students develop critical consciousness, learn to question assumptions, and become co-creators of knowledge, then AI tutoring is not merely insufficient but potentially counterproductive. Research from 2024--2025 documents what scholars term ``cognitive offloading'': students who rely heavily on LLMs for writing and research tasks demonstrate poorer reasoning and argumentation skills, focus on a narrower set of ideas, and produce more superficial analyses than those using traditional methods. AI risks automating and perfecting the banking model, producing students who can retrieve information efficiently but cannot think critically about it.

\textbf{Automated essay grading} illustrates the boundary precisely. AI can reliably grade essays on surface features that correlate with quality in standardized assessments. But evaluating originality of thought, depth of argumentation, creative risk-taking, and genuine intellectual engagement requires normative judgments about what constitutes good thinking. These are essentially contested concepts \citep{gallie1956}.

\textbf{Academic integrity} is adversarial co-evolution in its purest educational form. Every plagiarism detection tool triggers new evasion strategies. The fundamental question is not technical (can we detect AI-generated text?) but normative (what does ``learning'' mean when AI can produce the output?). This question has no algorithmic answer.

\textbf{Student advising and mentorship} are relationally irreducible. A faculty advisor who has worked with a graduate student for years possesses metis about that student that no AI system can acquire, because the knowledge is constituted in the relationship itself. \citet{biesta2009} argues that reducing education to measurable ``learning outcomes'' (what he calls ``learnification'') strips away the dimensions of education that matter most: socialization and subjectification (the development of independent agency).

\textbf{Admissions decisions} combine every Metis AI characteristic simultaneously. They are consequentially irreversible (an admissions decision shapes a life trajectory), relationally complex (holistic review), normatively open-textured (``merit'' is an essentially contested concept, as the \emph{SFFA v.\ Harvard} decision demonstrated), adversarial (applicants optimize against perceived criteria), and accountability-anchored (institutions bear legal and ethical responsibility).

\begin{table}[h]
\centering
\caption{Education tasks across the zone spectrum.}
\label{tab:education}
\small
\begin{tabular}{@{}lll@{}}
\toprule
Education task & Zone & Dominant pillars \\
\midrule
Vocabulary / grammar drills & 1 (AI-Native) & --- \\
Math tutoring (well-structured) & 1 (AI-Native) & --- \\
Standardized test preparation & 1 (AI-Native) & --- \\
Automated essay grading & 1 $\to$ 2 & Normative \\
Academic integrity enforcement & 2 (Boundary) & Adversarial, normative \\
Teaching critical thinking / ethics & 2 (Boundary) & Normative, relational \\
Student advising / mentorship & 2 (Boundary) & Relational, accountability \\
Curriculum design & 2 (Boundary) & Normative \\
Admissions decisions & 2 (Boundary) & All five pillars \\
\bottomrule
\end{tabular}
\end{table}

\subsection{Cross-Domain Pattern}

Across all four domains, a consistent pattern emerges. The boundary between Zone~1 and Zone~2 is not random; it tracks the transition from tasks dominated by \emph{uncertainty} (missing information that more data can supply) to tasks dominated by \emph{equivocality} (conflicting interpretations that only human judgment can negotiate). Within each domain:

\begin{enumerate}
    \item \textbf{Pattern recognition and optimization} are AI-native: reading images, scoring risk, classifying documents, drilling vocabulary.
    \item \textbf{The application of recognized patterns to consequential decisions in institutional contexts} is Metis AI: the same image read by the same algorithm requires a different response depending on who the patient is, what the legal standard requires, whether the counterparty is gaming the system, and who bears responsibility for getting it wrong.
\end{enumerate}

The five pillars are not equally distributed across domains. Finance is dominated by adversarial co-evolution and consequential irreversibility (reflexive markets, flash crashes). Healthcare is dominated by relational irreducibility and accountability anchoring (the physician--patient relationship, malpractice liability). Law is dominated by normative open texture and accountability anchoring (open-textured statutes, the requirement for judicial judgment). Education is dominated by normative open texture and relational irreducibility (contested concepts of merit and learning, the teacher--student relationship). But every domain, at its boundary, exhibits at least three of the five pillars simultaneously---confirming that the boundary zone is not a single-variable phenomenon but a structural feature of institutional life.

Figure~\ref{fig:heatmap} consolidates this pattern into a single view. Each row is a representative task drawn from the case studies and domain analyses; each column is a pillar. The shading reflects the intensity of that pillar's presence (\textcolor{red!80!black}{\rule{0.8em}{0.8em}}~high, \textcolor{orange!70}{\rule{0.8em}{0.8em}}~moderate, \textcolor{gray!20}{\rule{0.8em}{0.8em}}~low/absent). Two patterns are immediately visible: (1)~Zone~1 tasks have empty rows, while Zone~2 tasks activate three or more pillars; and (2)~certain pillar pairs---irreversibility with accountability, normative open texture with adversarial co-evolution---consistently co-occur, reflecting the reinforcing dynamics discussed in Section~\ref{sec:hitl}.

\begin{figure}[t]
\centering
\small
\setlength{\tabcolsep}{3pt}
\renewcommand{\arraystretch}{1.15}
\newcommand{\hi}{\cellcolor{red!35}}    
\newcommand{\md}{\cellcolor{orange!25}} 
\newcommand{\lo}{\cellcolor{gray!8}}    
\begin{tabular}{@{}l|l|c|c|c|c|c|c@{}}
\toprule
\textbf{Domain} & \textbf{Task} &
\rotatebox{70}{\textbf{Irreversibility}} &
\rotatebox{70}{\textbf{Relational}} &
\rotatebox{70}{\textbf{Normative}} &
\rotatebox{70}{\textbf{Adversarial}} &
\rotatebox{70}{\textbf{Accountability}} &
\rotatebox{70}{\textbf{Zone}} \\
\midrule
\multirow{3}{*}{\emph{Cases}}
& Content moderation     & \hi H & \md M & \hi H & \hi H & \md M & 2 \\
& Humanitarian allocation& \hi H & \md M & \hi H & \md M & \hi H & 2 \\
& Regulatory compliance  & \md M & \md M & \hi H & \md M & \hi H & 2 \\
\midrule
\multirow{4}{*}{\emph{Finance}}
& HF market-making       & \lo   & \lo   & \lo   & \lo   & \lo   & 1 \\
& Credit underwriting    & \md M & \lo   & \hi H & \lo   & \hi H & 2 \\
& Active fund mgmt (crisis) & \hi H & \lo & \lo  & \hi H & \md M & 2 \\
& AML/KYC compliance     & \md M & \lo   & \hi H & \hi H & \hi H & 2 \\
\midrule
\multirow{4}{*}{\emph{Health}}
& Radiology classification & \lo & \lo  & \lo   & \lo   & \lo   & 1 \\
& Treatment planning     & \md M & \hi H & \hi H & \lo  & \hi H & 2 \\
& End-of-life care       & \hi H & \hi H & \hi H & \lo  & \hi H & 2 \\
& Organ allocation       & \hi H & \md M & \hi H & \lo  & \hi H & 2 \\
\midrule
\multirow{4}{*}{\emph{Law}}
& E-discovery            & \lo   & \lo   & \lo   & \lo   & \lo   & 1 \\
& Sentencing / bail      & \hi H & \lo   & \hi H & \lo   & \hi H & 2 \\
& Trial strategy         & \md M & \hi H & \md M & \hi H & \md M & 2 \\
& Complex mediation      & \hi H & \hi H & \hi H & \md M & \md M & 2 \\
\midrule
\multirow{4}{*}{\emph{Educ.}}
& Math tutoring          & \lo   & \lo   & \lo   & \lo   & \lo   & 1 \\
& Academic integrity     & \lo   & \lo   & \hi H & \hi H & \md M & 2 \\
& Student advising       & \md M & \hi H & \md M & \lo   & \hi H & 2 \\
& Admissions decisions   & \hi H & \hi H & \hi H & \hi H & \hi H & 2 \\
\bottomrule
\end{tabular}
\caption{Task--pillar heatmap across domains. H~=~high, M~=~moderate. Shading intensity reflects pillar presence. Zone~1 rows are uniformly light; Zone~2 rows activate three or more pillars, with characteristic domain-specific clustering.}
\label{fig:heatmap}
\end{figure}

\section{Constitutive vs.\ Supervisory: Why Human-in-the-Loop Fails}
\label{sec:hitl}

A common response to Metis AI tasks is to propose ``human-in-the-loop'' (HITL) designs: post-hoc supervisory designs in which the AI produces a substantive recommendation and the human is asked to review, approve, or reject it. This sounds reasonable. The evidence suggests it fails.

Research on automation bias \citep{mosier2021} documents a persistent pattern: when AI systems provide recommendations, human overseers tend to accept them uncritically, even when the recommendations are wrong. This ``deep automation bias'' is not a matter of individual laziness but a structural consequence of the oversight architecture. The Verfassungsblog critique \citep{schwemer2025warm} names this precisely: a ``warm body in the loop'' that provides the appearance of human control without its substance.

The problem is not bad implementation. The problem is a category error. HITL designs treat human involvement as \emph{supervisory}: a quality-control layer applied after the AI has done the substantive work. But in Metis AI tasks, human involvement is not only supervisory; it is \emph{constitutive}. The human is not only reviewing the AI's output; the human is \emph{performing the task}: exercising judgment, managing relationships, interpreting norms, bearing accountability, with AI as analytical support. 

\citet{garfinkel1967} showed that accountability is constitutive of social action: people make their actions ``accountable'' in the doing of them, not after the fact. \citet{ostrom1990} showed that effective governance requires monitors who are embedded in the community, not external oversight systems. \citet{habermas1981} showed that communicative legitimacy requires genuine participation, not post-hoc endorsement.

These theories converge on the same point: in legitimacy-sensitive social tasks, accountability, interpretation, and participation are not external checks that can be added after a decision has been generated. They are part of the activity through which the decision becomes socially valid. Thus, the higher the Metis content of a task, the less meaningful post-hoc review becomes, because determining whether the AI’s recommendation is appropriate requires the human to re-perform the very judgment that the system has already framed as completed. The reviewer is therefore asked not merely to check an output, but to overturn an entire decision frame under conditions of asymmetrical information and institutional pressure.

More specifically, the architectural distinction is between:
\begin{itemize}
    \item \textbf{Human-in-the-loop} (supervisory): AI acts, human reviews. Appropriate for Zone~1 tasks with occasional edge cases.
    \item \textbf{Human-in-the-lead} (constitutive): This is what really happening for Zone~2 tasks: Human acts, AI supports. The human is the decision-maker, the relational agent, the norm-interpreter, and the accountable party. AI provides data analysis, pattern detection, scenario modeling, and draft generation. This is what we called the ''centaur'' model.
\end{itemize}

For Metis AI tasks, the centaur is not a compromise between automation and human control. It is the architecture that best matches the structure of the task.

\section{Conclusion: The Boundary Shifts, but the Zone Persists}
\label{sec:conclusion}

With the advancement of technology, individual tasks can transition between zones. Face recognition moved from Zone~3 (embodied perception) to Zone~1 (AI-native digital classification). Simple content moderation is moving from Zone~2 toward Zone~1. Simple tax preparation has largely moved from Zone~2 to Zone~1. In each case, the task became AI-native when its essential difficulty was reducible to pattern-matching against well-labeled data.

But the Zone~2 category is permanent. This is because the five characteristics that define it: consequential irreversibility, relational irreducibility, normative open texture, adversarial co-evolution, and accountability anchoring, are not properties of current AI systems but properties of human institutional life. As long as human societies produce irreversible consequences, maintain interpersonal relationships, interpret contested norms, engage in strategic conflict, and require personal accountability, there will be digital tasks that demand metis rather than techne.

Moreover, the zone \emph{expands} as AI automates simpler tasks. \citet{autor2024} documented that approximately 60\% of employment in 2020 is in occupations that did not exist in 1940. As AI commoditizes existing digital tasks, human work shifts toward tasks that are harder to automate: tasks with more irreversibility, more relational complexity, more normative ambiguity, more adversarial pressure, and more accountability requirements. The boundary moves, but it moves \emph{with} us.

The implications of Metis AI are threefold.

\textbf{For AI system designers:} Use the five-pillar framework as a diagnostic before choosing an architecture. If a task scores high on three or more pillars, design for human-in-the-lead, not human-in-the-loop. Build centaur systems where AI provides analytical power and the human provides judgment, relational capacity, and accountability.

\textbf{For organizations deploying AI:} Resist the temptation to automate Metis AI tasks simply because they are digital. The fact that a task can be performed on a computer does not mean it can be delegated to one. Invest in the human capabilities: judgment, relational skill, normative interpretation, that Metis AI tasks demand, rather than assuming these will be automated away.

\textbf{For researchers:} The most important open questions about AI capability are not in Zone~1 (where benchmarks are mature) or Zone~3 (where the limitations are physical). They are in Zone~2, where the limitations are institutional, social, and normative. We need theoretical frameworks, empirical methods, and design principles for the Metis AI zone. The value lies in not (only) to automate it, but to augment the humans who work within it.

\bibliographystyle{plainnat}
\bibliography{references}

\end{document}